\documentclass{article} % For LaTeX2e
\usepackage{iclr2026_conference,times}

\usepackage{amsmath,amsfonts,bm}

\def\eqref#1{equation~\ref{#1}}
\def\1{\bm{1}}

\def\vf{{\bm{f}}}

\def\vh{{\bm{h}}}

\def\vo{{\bm{o}}}

\def\vx{{\bm{x}}}

\def\mF{{\bm{F}}}

\def\mH{{\bm{H}}}

\def\mO{{\bm{O}}}

\def\mX{{\bm{X}}}

\DeclareMathAlphabet{\mathsfit}{\encodingdefault}{\sfdefault}{m}{sl}
\SetMathAlphabet{\mathsfit}{bold}{\encodingdefault}{\sfdefault}{bx}{n}

\def\sR{{\mathbb{R}}}

\newcommand{\mymethod}{{\texttt{Secondary Sinks}}}

\usepackage[utf8]{inputenc} % allow utf-8 input
\usepackage[T1]{fontenc}    % use 8-bit T1 fonts
\usepackage{hyperref}       % hyperlinks
\usepackage{url}            % simple URL typesetting
\usepackage{booktabs}       % professional-quality tables
\usepackage{amsfonts}       % blackboard math symbols
\usepackage{nicefrac}       % compact symbols for 1/2, etc.
\usepackage{microtype}      % microtypography
\usepackage{multirow}

\usepackage[dvipsnames]{xcolor}
\usepackage{graphicx}
\usepackage[capitalize,noabbrev]{cleveref}
\usepackage{caption}
\usepackage{subcaption}
\usepackage{wrapfig}
\usepackage{amsthm}
\usepackage{makecell}
\usepackage{float}
\usepackage{soul}
\usepackage[most]{tcolorbox}
\definecolor{qkred}{HTML}{C5172E}
\definecolor{vogreen}{HTML}{118B50} %rgb(18, 158, 90)
\definecolor{ffnblue}{HTML}{1B56FD}

\definecolor{boxblue}{HTML}{edf2fb} % #e6ebf7
\newtcolorbox{mybox}{colback=boxblue,colframe=black}

\makeatletter
\newcommand{\DrawBoxLine}{%
  \begin{tikzpicture}
  \path[use as bounding box] (0,0) -- (\linewidth,0);
  \draw[color=black,dashed,dash phase=2pt]
        (0-\kvtcb@leftlower-\kvtcb@boxsep,0)--
        (\linewidth+\kvtcb@rightlower+\kvtcb@boxsep,0);
  \end{tikzpicture}%
  }
\makeatother

\title{On the Existence and Behavior of Secondary Attention Sinks}

\author{Jeffrey T.H. Wong\thanks{Part of this work was carried out during an internship at Unlikely AI.}$^{*1}$, Cheng Zhang$^{1}$, Louis Mahon$^{2}$, Wayne Luk$^{1}$, Anton Isopoussu$^{2}$, Yiren Zhao$^{1}$\\
$^{1}$Imperial College London \quad $^{2}$UnlikelyAI \\
\texttt{\{tsz.wong20,cheng.zhang122,w.luk,a.zhao\}@imperial.ac.uk} \\ \texttt{\{louis,anton\}@unlikely.ai}}

\iclrfinalcopy % Uncomment for camera-ready version, but NOT for submission.
\begin{document}

\maketitle
\lhead{Accepted to ICLR 2026 Workshop on Unifying Concept Representation Learning}
\theoremstyle{plain}
\newtheorem{theorem}{Theorem}
\newtheorem{corollary}{Corollary}[theorem]
\newtheorem{lemma}[theorem]{Lemma}
\newtheorem{remark}{Remark}
\theoremstyle{definition}
\newtheorem{assumption}{Assumption}
\newtheorem{definition}{Definition}
\newtheorem{problem}{Problem} % not sure if "Problem" is a proper keyword

\begin{abstract}
Attention sinks are tokens, often the beginning-of-sequence (BOS) token, that receive disproportionately high attention despite limited semantic relevance. In this work, we identify a class of attention sinks, which we term secondary sinks, that differ fundamentally from the sinks studied in prior works, which we term primary sinks. 
While prior works have identified that tokens other than BOS can sometimes become sinks, they were found to exhibit properties analogous to the BOS token. Specifically, they emerge at the same layer, persist throughout the network and draw a large amount of attention mass. Whereas, we find the existence of secondary sinks that arise primarily in middle layers and can persist for a variable number of layers, and draw a smaller, but still significant, amount of attention mass.
Through extensive experiments across 11 model families, we analyze where these secondary sinks appear, their properties, how they are formed, and their impact on the attention mechanism. 
Specifically, we show that:
(1) these sinks are formed by specific middle-layer MLP modules; these MLPs map token representations to vectors that align with the direction of the primary sink of that layer,
(2) the $\ell_2$-norm of these vectors determines the sink score of the secondary sink, and also the number of layers it lasts for, thereby leading to different impacts on the attention mechanisms accordingly,
(3) the primary sink weakens in the middle layers, coinciding with the emergence of secondary sinks.
We observe that in larger-scale models, the location and lifetime of the sinks, together referred to as sink levels, appear in a more deterministic and frequent manner. Specifically, we identify three sink levels in QwQ-32B and six levels in Qwen3-14B. 
We open-sourced our findings at \href{https://github.com/JeffreyWong20/Secondary-Attention-Sinks?tab=readme-ov-file}{\color{Blue}{github.com/JeffreyWong20/Secondary-Attention-Sinks}}.

\end{abstract}

\section{Introduction}

Attention sinks were first identified by~\citet{first-attn-sink}, where the BOS token was observed to receive anomalously high attention weights. This phenomenon has since shown broad practical implications, including LLM quantization~\citep{prefixing-attn-sink, liu2024intactkv}, KV-cache optimization~\citep{rkv, cai2024pyramidkv}, efficient LLM serving~\citep{first-attn-sink}, and model  enhancement~\citep{calibration-sink}.

Many recent studies investigated the formation and functional role of the BOS sink. \citet{cancedda2024spectral} analyzes attention sinks from a spectral subspace perspective, while \citet{When-attention-sink-emerges} interprets them as a form of positional bias that mitigates over-mixing. Building on this line of work, \citet{two-sides-of-the-same-coin} further examines their role in the depth-wise organization of the model, proposing that attention sinks serve as a mechanism for information compression along the depth dimension.

\citet{massive-activation} and \citet{calibration-sink} show that attention sinks are not limited to the BOS token; instead, multiple tokens can function as attention sinks. \citet{What-are-you-sinking} further analyzes this phenomenon from a geometric perspective, showing that the emergence of multiple attention sinks is closely tied to the model’s positional embedding scheme. In this view, attention sinks act as reference points in a high-dimensional representation space, enabling the model to establish a stable internal coordinate system. However, the multiple sinks identified in previous work are fundamentally the same as the BOS sink: they emerge at the same layers and persist throughout the network. In contrast, we identify a new type of sink that differs in both its layer of emergence and its lifetime.

% the understanding of the attention mechanism remains limited, particularly with respect to the role of multiple attention sinks that emerge during generation and how they differ from the BOS sink.

In this work, we show that all of the multiple tokens that act as attention sinks can be organized into distinct \textit{sink levels}. The primary level normally corresponds to the BOS sink: it emerges at the same layer as the BOS token and persists throughout the network. Additional sink levels arise in the middle layers and persist for a variable number of layers; we refer to these as \mymethod{}. 
% Subsequent sections study these secondary sinks in detail. 

%We first characterize the similarities and differences between secondary sinks and the BOS sink across a range of models, 
%identifying the token sets and position in which they most frequently occur (\Cref{section:3}). 

We first identify \mymethod{}, especially their \textit{sink levels}, the token sets and positions in which they frequently occur, through
characterizing the similarities and differences between \mymethod{} and the BOS sink across a range of models. 
% We then analysis the \textit{sink levels} distribution across different model scales.
(\Cref{section:3}).
%We then conduct an empirical analysis of their emergence across network depth, quantifying the contributions of different layers to their formation (\Cref{section:4}). 
We then quantify the contributions of different layers to their formation via an empirical analysis of their emergence across network depth (\Cref{section:4}).
Finally, we examine their impact on attention scores throughout the network.(\Cref{section:5}) . We make the following conclusions:

% \begin{itemize}
%     \item Unlike BOS sink that occurs in early layer and span through the entire network, \mymethod{} occurs mainly in middle layer span through various number of layers.
%     \item \mymethod{} share similar direction as BOS Sink. This direction is encoded in specific \textbf{middle MLP module} in the networks and can be recovered by multiple orthogonal directions.
%     \item \mymethod{} created at different layers shows different levels of life time. The life-time of these sinks are highly correlated to the norm of the residual stream with larger model showing more consistent distribution across layers and life-time.
%     \item BOS Attention sinks decades and reaches the weakest points in the middle layers while secondary Sinks starts occurring in the middle layer.
%     \item This secondary sinks occur deep in the generated sequence.
% \end{itemize}

\begin{itemize}
    \item Unlike the primary sink, which emerges in early layers and persists throughout the entire network, \mymethod{} arise primarily in middle layers and persist only for a couple of layers. 
    They can be found at any position in the generation sequence and semantically uninformative tokens.

    \item \mymethod{} shares a similar direction with the primary sink. This direction is encoded in specific \textbf{middle-layer $l_{\text{start}}$ MLP modules}, which convert multiple orthogonal directions to the same sink direction. After $l_{\text{start}}$, a set of semantically uninformative tokens are transformed into attention sinks. The layers preceding $l_{\text{start}}$ play a key role in constructing this set, distinguishing these tokens from other semantically uninformative tokens.

    \item Different levels of \mymethod{} exhibit distinct lifetimes and attention sink strength. Both are strongly correlated with the $\ell_2$-norm of $l_{start}$ output. Larger models show clearer differentiation between sink levels, and models going through extensive post-training on reasoning data show stronger \mymethod{} phenomenon.
    
    \item \mymethod{} show a compensating effect relative to the BOS sink. The BOS gradually decays and reaches its weakest strength in the middle layers, coinciding with the emergence of \mymethod{} phenomenon.

\end{itemize}

\section{Background and Preliminary}

Let $f_{\theta}$ be a decoder-only transformer with $L$ layers and hidden size $h$.
At each decoder layer $l$, the decoder receives a hidden sequence of length t, $\mH^l \in \sR^{t \times h} = \{ \vh_0^l, \vh_1^l, \ldots, \vh_t^l\}^T$, 
% \cz{does this $\{...\}$ represent a matrix? if so a transpose is needed?}
where $\vh_i$ is the hidden representation at position $i$ in layer $l$.

% \paragraph{Decoder blocks} Each decoder layer $l$ consists of an multi-head self-attention module (MHSA) and a multi-layer perceptron (MLP). Both modules read from and write to the residual stream which is equivalent to the hidden state input at that layer, $H^l \in \sR^{t \times h}$ that connects the entire network. We denote the output of MHSA as $O^l \in \sR^{t \times h}$ and MLP as $F^l \in \sR^{t \times h}$. Depending on the architecture, the decoder may use either pre-norm or post-norm, which determines whether layer normalization is applied before or after each module. 

\paragraph{Decoder blocks} Each decoder layer $l$ consists of a multi-head self-attention (MHSA) module and a multi-layer perceptron (MLP)
% as shown in~\Cref{fig:decoder overview and cosine simiarity within mlp}
% \cz{this cref should be Fig}
, both of which operate on the decoder input hidden state, also referred to as the residual stream $\mH^l \in \sR^{t \times h}$. The MHSA produces an output $\mO^l \in \sR^{t \times h}$, and the MLP produces $\mF^l \in \sR^{t \times h}$; in both cases, the outputs are added back to the residual stream. Decoders may use either a pre-norm or post-norm architecture for which the normalization is applied either before or after each module. The majority of modern models employ pre-norm as shown in~\Cref{fig:decoder overview and cosine simiarity within mlp}.
% \cz{this too}.

\paragraph{Position Embedding} 
% Position embeddings $P$ are used in the attention module to provide the model with positional information for each token in the sequence. There are several types of positional embeddings, including absolute positional embeddings~\citep{attention-is-all-you-need}, rotary embeddings (RoPE)~\citep{rope}, and NTK\footnote{NTK stands for Neural Tangent Kernel, this refers to changing the base frequency in RoPE based on the network capability, allowing improved performance on long-context sequences.}-aware scaled RoPE~\cite{yarn, effective-long}. 
Position embeddings $P$ provide tokens with positional information in the attention mechanism. Common $P$ include absolute positional embeddings~\citep{attention-is-all-you-need}, rotary positional embeddings (RoPE)~\citep{rope}, and NTK\footnote{NTK stands for Neural Tangent Kernel, this refers to changing the base frequency in RoPE based on the network capability, allowing improved performance on long-context sequences.}-aware scaled RoPE~\citep{yarn, effective-long}.
% Among these, RoPE and NTK-aware scaled RoPE are the most commonly used in modern large language models. RoPE is widely adopted by architectures such as LLaMA~\cite{llama2}, Mistral~\cite{mistral7b}, while NTK-aware scaled RoPE is used in models such as Qwen~\cite{team2024qwen2}, CodeLlama~\cite{Codellama}. 
RoPE is used in models such as LLaMA~\citep{llama2} and Mistral~\citep{mistral7b}, while NTK-aware scaled RoPE is adopted by Qwen~\citep{team2024qwen2} and CodeLlama~\citep{Codellama}. Compared to RoPE, NTK-aware scaled RoPE increases the rotary base resulting in less rotation with position~\citep{effective-long}.
\paragraph{Attention Sink} Attention sink refers to the phenomenon where semantically uninformative tokens are assigned disproportionately high attention weights across diverse model architectures and scales, resulting in vertical patterns in the attention weight matrix~\citep{first-attn-sink}.

% In this work, we show that although multiple tokens can act as attention sinks, they can be organized into distinct \textit{levels} of sinks. The first level corresponds to the BOS sink: it emerges in the same layer as the BOS token and persists throughout the network. Additional levels arise in the middle layers and persist for a variable number of layers; we refer to these as \mymethod{}. Subsequent chapters study these distributed sinks in detail.

% We observe that models (Qwen2,Qwen2.5,Qwen3,QwQ) with 7B parameters and above exhibit at least 3 sink levels, while Qwen3-14B shows 6 distinct levels. These levels differ substantially across levels in norm, with each level exhibiting an order-of-magnitude difference relative to others, while still remaining several orders of magnitude stronger than typical tokens.

Recent work~\cite{What-are-you-sinking} suggests that reducing the rotary frequency weakens the dot-product advantage of the first token. As a result, standard RoPE often produces a strong attention sink at the BOS token, whereas NTK-aware scaled RoPE can give rise to additional attention sinks at other positions in the sequence. In this work, we evaluate 11 model families that use distinct rotary base frequencies (10K-500K). Our findings on QwQ and the Qwen2/2.5/3 model families aligns with~\cite{What-are-you-sinking}. However, we observe that CodeLlama and several other model families, despite using very large rotary bases, do not exhibit such secondary sinks. The underlying cause of this discrepancy remains an open question.

In this work, we show that while multiple tokens can act as attention sinks, they can be organised into discrete levels. The primary level corresponds to the BOS sink: these sinks emerge in the same layer as the BOS token and persist throughout the network. Additional levels appear in later layers and persist for varying depths. 
Building on this, we study the properties of additional sink levels, the variation of this phenomenon across different model sizes and post-training methods, then examine their formation across network depth, and finally analyze their resulting impact on attention.
% Finally, we quantify the variation of this phenomenon across different model sizes and post-training methods.

% \begin{itemize}
%     \item LM models
%     \item Introduce transformer architecture and what is attention.
%     \item Positional Embedding
%     \item Attention Sinks
% \end{itemize}

\section{Properties of \mymethod{}}
\label{section:3}

\begin{figure}[ht]
    \centering
    \begin{subfigure}[t]{0.85\linewidth}
        \centering
        \includegraphics[width=\linewidth]{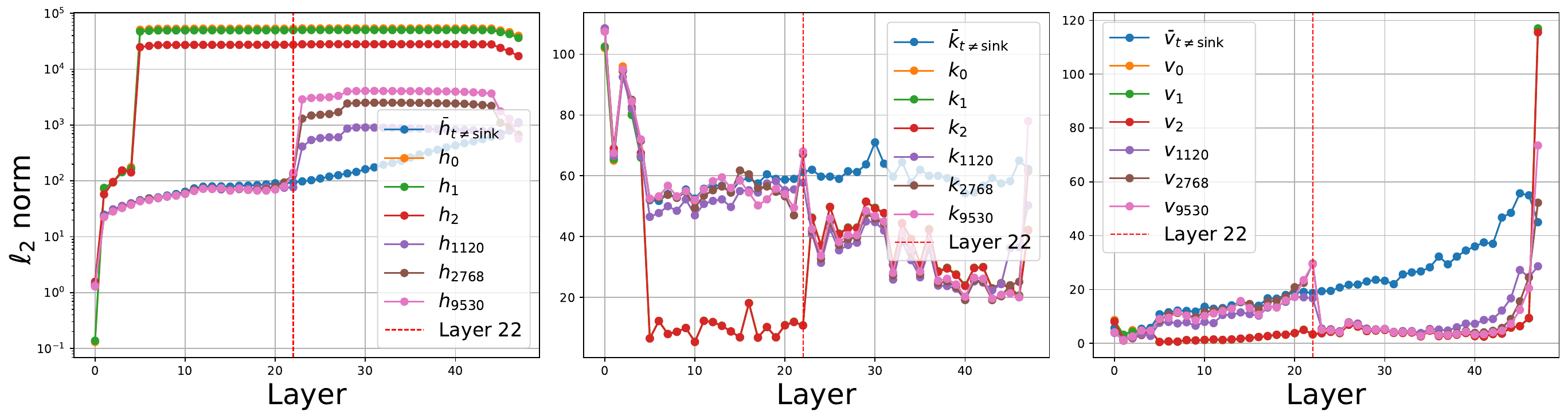}
        % \caption{DeepSeek-14B}
        \label{fig:deepseek}
    \end{subfigure}

    % \vspace{0.5em}

    \begin{subfigure}[t]{0.85\linewidth}
        \centering
        \includegraphics[width=\linewidth]{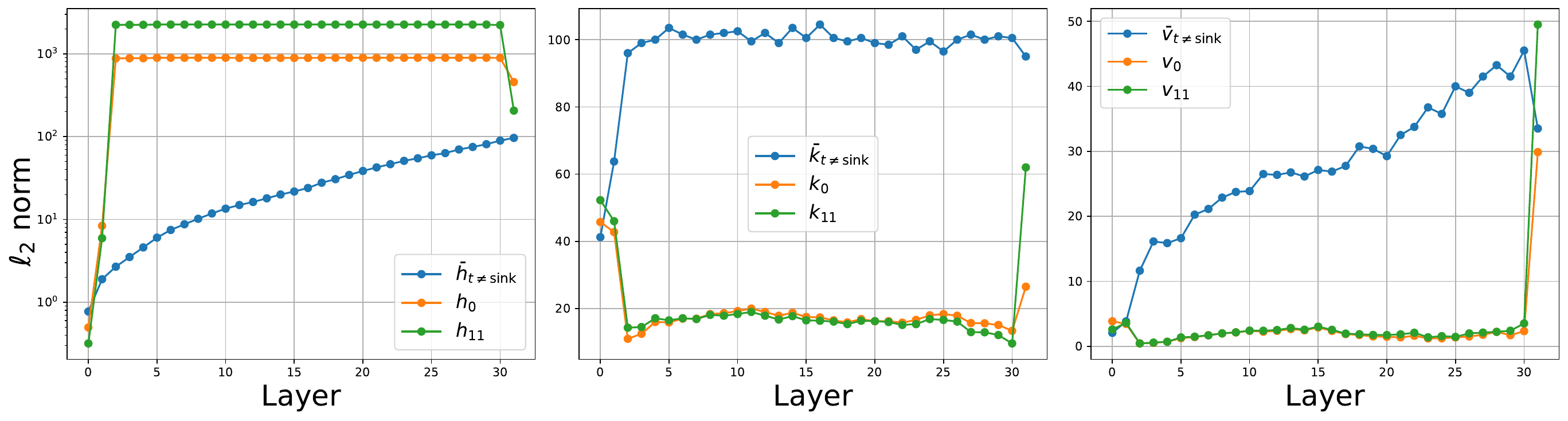}
        % \caption{LLaMA2-7B-Chat}
        \label{fig:llama}
    \end{subfigure}
    \vspace{-5pt}
    \caption{\textbf{(Bottom)} In LLaMA2-7B-Chat, the 1st, 11th tokens act as primary sinks: they exhibit significantly larger $l_2$ norms in their hidden states, while having much smaller key and value norms compared to other tokens. \textbf{(Top)} DeepSeek-14B also shows primary sinks (tokens 0, 1, and 2). In addition, it exhibits other attention sinks that emerge in middle layers (e.g., tokens 1120, 2768).}

    \label{fig:overview}
\end{figure}

\vspace{-5pt}
We perform extensive empirical experiments to study the properties of \mymethod{} by comparing them with the BOS sink. To observe the \mymethod{}, we generate reasoning traces from LLMs (DeepSeek-14B, DeepSeek-32B, Qwen-3-32B) on datasets including AIME24~\citep{AIME2024}, Math~\citep{math-500}, and then threshold the attention weights to locate sink tokens across different layers. We find that these attention sinks consistently exhibit at least an order of magnitude higher $\ell_2$-norms than average token\footnote{Tokens excluding attention sinks.}
% \cz{do we need to define average token here?}}
and align along the similar the BOS sink direction.

Based on this observation, we developed an efficient method to identify attention sinks by computing the pairwise cosine similarity between each hidden state \textcolor{purple}{$\vh_{t!=0}^l$} and the BOS token \textcolor{orange}{$\vh_0^l$} at layer $l$, and then thresholding with cosine similarity $>$ 0.95. 
We also observe that this phenomenon can be triggered by naturally distributed inputs rather than model-generated inputs. Accordingly, we use reasoning traces generated by DeepSeek-14B on the two datasets and feed them into 11 model families: Deepseek-distill, Qwen2, Qwen2-Math, Qwen2.5, Qwen2.5-Math, QwQ, Qwen3, LlaMA-3.1, Phi-4, Mathtral, CodeLlama. In the main body of the paper, we focus on results from DeepSeek-14B on AIME24; we observe the same behaviour in other models that exhibit secondary sinks (See Appendix~\cref{apx:sec:detection results}).

% \cz{we can use colored symbols like \textcolor{Green}{$h_1$} for ease of read}
\paragraph{Secondary Sinks} \Cref{fig:overview} (top) shows the $\ell_2$-norms of the hidden states, keys, and values of attention sinks, compared with those of an average token in DeepSeek-14B denoted by \textcolor{blue}{$\bar{\vh}_{t\neq \text{sink}}$}. Token \textcolor{orange}{$\vh_0$} corresponds to the BOS token, while tokens \textcolor{orange}{$\vh_1$} and \textcolor{orange}{$\vh_2$} are tokens from the chat template. We classify all of these sinks as \textit{primary sinks}, as they emerge simultaneously and persist for the same number of layers as the BOS sink.
In addition to these BOS sinks, we observe \textit{secondary sinks} at positions 1120, 2768, and 9530. 
These tokens behave like normal tokens up to layer 22, after which they are converted into sinks. 
Before layer 22, their hidden-state, key, and value norms follow the average-token trend; afterward, they closely track the BOS sink, whose hidden-state norm is several orders of magnitude higher than that of normal tokens, while its key and value norms are lower.
% \cz{highlight layer 22 in figures}

Not all models exhibit secondary sinks. For example, although multiple sinks also appear in LLaMA-7B-Chat~\Cref{fig:overview} (Bottom), they are exclusively primary sinks. Among the 11 model families we analyzed, secondary sinks were found in Qwen2, Qwen2-Math, Qwen2.5, Qwen2.5-Math, Qwen3, and QwQ. 
Full detection results on all 11 model families can be found in Appendix~\ref{apx:sec:detection results}. These sinks typically occur in semantically uninformative tokens~\Cref{tab:token-distribution} and 
can appear at any position during the generation~\Cref{fig:token-pos-density}.
% are more likely to appear at deeper sequence positions~\Cref{fig:token-pos-density}.

\begin{figure}[h!]
    \centering
    \includegraphics[width=0.8\linewidth]{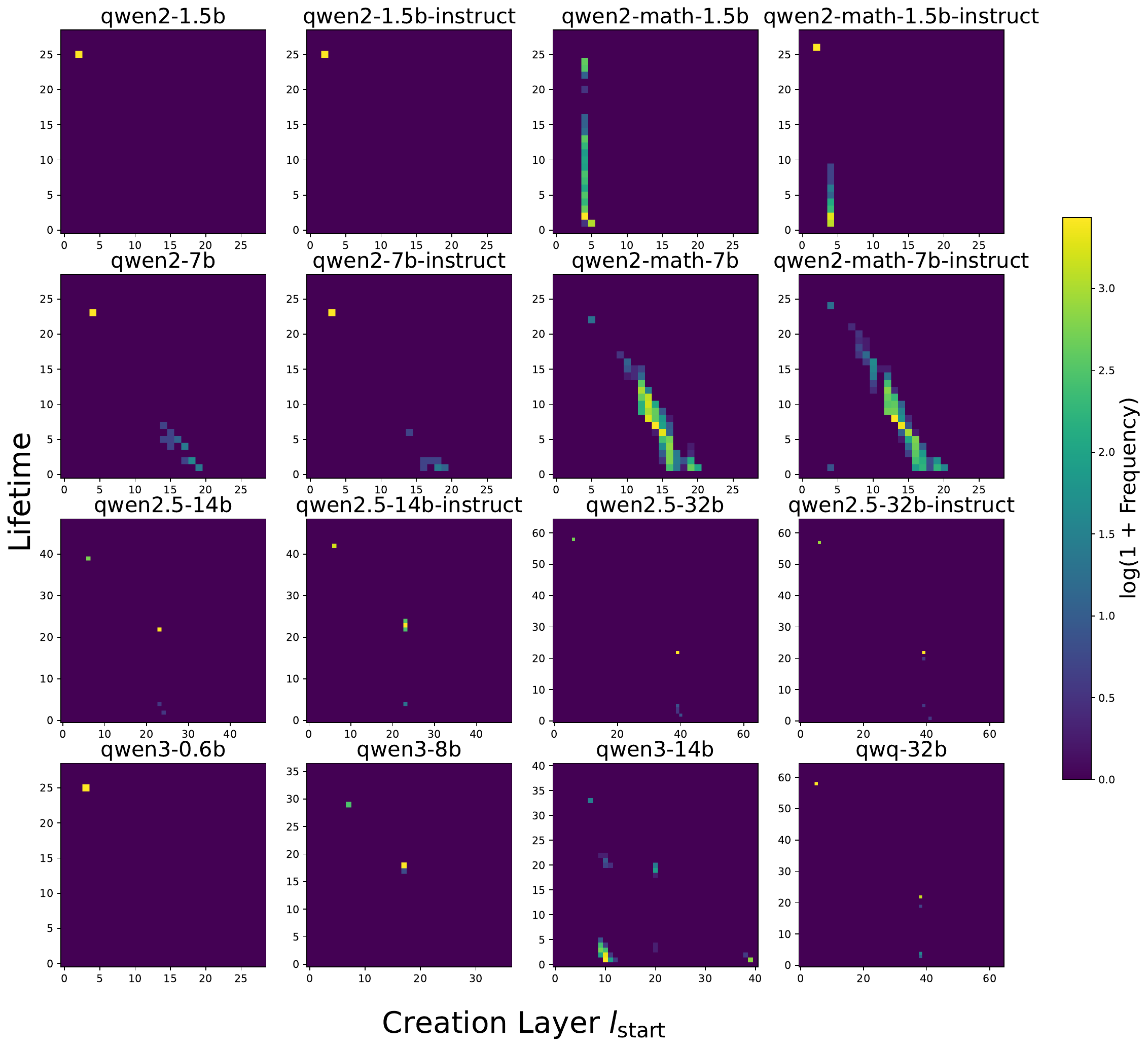}
    \caption{Sink levels ($l_\text{start}$, \textit{lifetime}) at different model scales.}
    \label{fig:sink levels}
\end{figure}

\vspace{-5pt}
\paragraph{Sink Levels} 
% \Cref{fig:lifetime} shows that secondary sink can persist for a varying number of layers after its creation. 
As well as differentiating primary from secondary sinks, we can, to some extent, go further and group secondary sinks into multiple `levels', based on the layer they begin in and how many layers they persist. 
%Secondary sinks can persist for a varying number of layers after its creation at $l_{\text{start}}$. after its creation at $l_{\text{start}}$. 
We associate each sink with an attribute pair $(l_{\text{start}}, \text{\textit{lifetime}})$, and refer to sinks with different attribute pairs as belonging to different sink levels. We quantify the distribution of sink levels across models of different sizes and families, as shown in \Cref{fig:sink levels}. 
% \cz{if we run out of pages, we can only keep 4 subfigures and move others to appendix}

Secondary sinks are absent or very weak in small base models. They begin to emerge more prominently after mid-training on large amounts of mathematical data, as evidenced by the transition from Qwen2-1.5B/7B to Qwen2-Math-1.5B/7B. This suggests that secondary sinks arise as a mechanism to support enhanced reasoning capability.
As model scale increases further, we observe that in the majority of models the number of distinct sink levels decreases: sink lifetimes and creation layers become less dispersed and increasingly concentrated at a small number of characteristic levels across the network.

% \input{figures/fig_sink_levels_model_scale}

% \begin{mybox}
%     \textbf{Finding 1.1: Middle Layer Emergence} \hfill $\triangleright$ \Cref{fig:overview}
%     \vspace{0.5em}

%     Unlike the typical BOS sink tokens that appear in the early layers of the network,
%     secondary sinks tend to emerge in the middle layers of LLMs. 
%     \DrawBoxLine
%     \vspace{0.5em}
%     \textbf{Finding 1.2: Model Size Effect on Sink Levels} \hfill $\triangleright$ \Cref{fig:sink levels}
%     \vspace{0.5em}

%     Larger models tend to exhibit fewer but more consistent and clearly differentiated levels.

% \end{mybox}

% \input{figures/fig_token_pos_density}
\section{Causal Formation Of \mymethod{}}
\label{section:4}
% \cz{I'm not sure about this section. Please help to correct it.}

% Outline:
%
% \begin{itemize}
%     \item We first analysis where these sinks come from. We show that the primarily comes from specific MLP layers in the network.
%     \item Through the MLP layers the cos-similarity get closer and closer to 1.
%     \item To identify the casual formation, we conducted clustering on early layers between normal uninformative tokens and uninformative tokens that got turned into a sink.
%     \item We perform activation swapping between average tokens and sinks tokens in early layers with its residual stream, mlp and attention. The results shows that its effect can be trace back to very earlier layers.
% \end{itemize}

% The secondary sinks exhibit intriguing formation patterns within the model layers. As shown in~\cref{fig:overview}, the secondary sinks are only converted into sinks after layer 22, before that all its norm are similar to normal tokens.
% To investigate their origins,
% we zoom in layer 22 and calculate the cosine similarity between tokens that eventually become secondary sinks and the BOS sink token within the layer,
% and find that these sinks primarily emerge from specific MLP layers in the network.
% We additionally perform clustering analysis on early layers,
% and token swapping experiments to trace back the formation of secondary sinks to very early layers of the model.

The secondary sinks exhibit an intriguing formation patterns across model layers. As shown in~\cref{fig:overview}, they only become apparent as sinks after layer 22; before this layer, their hidden state $\ell_2$-norms are similar to those of normal tokens. To investigate their origins, we track tokens that eventually become secondary sinks, and compute their cosine similarity with the BOS sink token as they pass through the different components of layer 22. 
Here we focus on presenting results on Deepseek-14B on AIME24. Results on other models and datasets can be found in~\cref{apx:sec:detection results}.
Our analysis reveals that these sinks primarily emerge from specific MLP modules in the network.
Additionally, we perform clustering analysis on early layers and conduct token swapping experiments to trace the formation these sinks back to very early layers of the model.

% \begin{figure}[h!]
%     \centering
%     \includegraphics[width=0.6\linewidth]{figures/raws/mlp_trace.pdf}
%     \caption{Cosine Similarity Between the Different Distributed Sinks and BOS Sink Token as they go through the Layer 22 in Deepseek-14B.}
%     \label{fig:token-pos-density}
% \end{figure}

\vspace{-15pt}
\begin{figure}[h!]
    \centering
    \begin{subfigure}{0.25\linewidth}
        \centering
        \includegraphics[width=\linewidth]{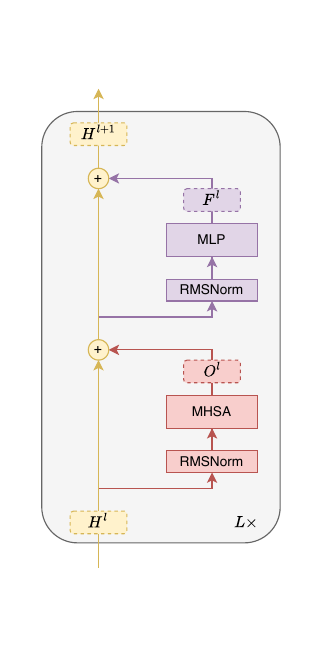}
        % \caption{Decoder overview}
        \vspace{-9mm}
        % \label{fig:decoder overview}
    \end{subfigure}
    % \hfill
    \begin{subfigure}{0.6\linewidth}
        \centering
        \includegraphics[width=\linewidth]{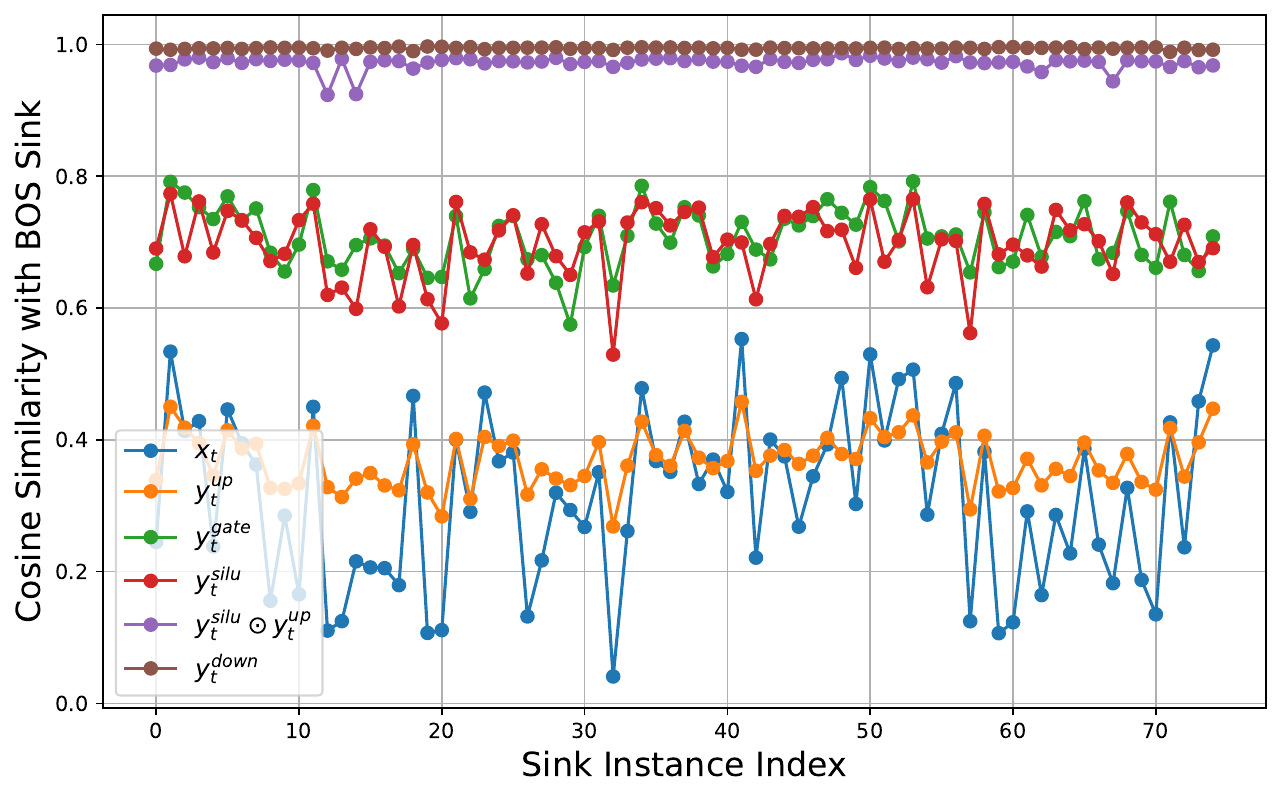}
        % \caption{Cosine Similarity Between the Different Distributed Sinks and BOS Sink Token as they go through the Layer 22 in Deepseek-14B.}
        % \label{fig:cosine simiarity within mlp}
    \end{subfigure}
    \caption{\textbf{(Left)} Decoder overview \textbf{(Right)} Cosine Similarity Between the Different Secondary Sinks and BOS Sink Token as they go through the Layer 22 MLP in Deepseek-14B. 
    % \cz{add axis labels like cos similarity and layer idx?}
    }
    \label{fig:decoder overview and cosine simiarity within mlp}
\end{figure}

% \input{figures/fig_mlp_pca_analysis}
% \cz{Correct/Add experiment details}

\paragraph{Cosine Similarity \& PCA Analysis} For each thinking trace,
we compute the cosine similarity between the hidden vector that eventually becomes a secondary sink
and that of the BOS sink token at each layer.
For simplicity, we name the tokens that eventually become secondary sinks as "future secondary sinks".

\Cref{fig:decoder overview and cosine simiarity within mlp} (Right) shows the cosine similarity of 75 future secondary sink tokens from AIME24 in DeepSeek-14B (layer 22). The MLP inputs $\vx_t$ initially have low similarity to the BOS direction, but this similarity increases through the MLP; by the output $\vf_t$, the vectors are nearly fully aligned with the BOS sink.

This suggests that the sink direction is encoded in the MLP and that future sink tokens contain enough information to recover it. Following the linear hypothesis~\citep{park2023linear}, we hypothesize that these tokens share a common key that retrieves the sink direction. We test this by performing PCA on the concatenated MLP inputs $\mX^l$ from 250 sink tokens. The spectrum is strongly low-rank. Feeding the top individual principal components through the MLP shows that these PCs are mapped to the same sink direction. \Cref{fig:pca} plots the MLP output norm and cosine similarity with the BOS sink when feeding $\pm\alpha \mathrm{PC}_i$. We find that the MLP amplifies components aligned with the BOS sink and suppresses those that are misaligned, producing the large hidden-state norms.
% ================== more concise version ================
% \input{figures/fig_tsne}

\paragraph{Clustering}
After identifying the layers at which secondary sinks begin to emerge, we conduct a clustering analysis on the hidden states $\vh_t^l$, attention outputs $\vo_t^l$, and MLP outputs $\vf_t^l$. These vectors correspond to two categories: (1) normal uninformative tokens and (2) future secondary sinks.

As shown in \Cref{fig:clustering}, in early layers, normal uninformative tokens and future sinks do not form clearly separable clusters. As the network depth increases, these two categories gradually become distinguishable, forming two increasingly distinct clusters. Both the attention and MLP modules contribute to representations that separate the two types of tokens. Notably, we observe the emergence of two well-defined clusters as early as layer 19. This shows that, although the \emph{effect} of the sink doesn't emerge until layer 22, the \emph{decision to create the sink} begins in earlier layers and is mostly complete before layer 22 itself. 

To better isolate the decision-making contributions from attention, MLP, and the hidden state, we substitute their outputs at future sink tokens with the average semantically uninformative token. See Appendix \ref{token-swapping} for further details.

\section{\mymethod{} Effect on Attention}
\label{section:5}
% Outline:
% \begin{itemize}
%     \item Interestingly we observed that BOS sink effect follows a valley shape across the model layers. When BOS sinks reaches the bottom of the valleys, \mymethod{} started emerging.
%     \item \mymethod{} can take away different amount of attention sinks from 1\% to 10\% and they can last for various of layers. We found out that their life time and sinks rate are both correlated with the norm of MLP output.  We term this secondary levels.
%     \item The secondary levels are related to the size of the model with larger size models following showing fewer and more consistence secondary levels.
% \end{itemize}

% Inspired by the previous works on the BOS sink, we further investigate how secondary sinks affect attention score distribution
% and the correlation between the norm of secondary sinks and sink rate and its life time~\addref{}.

% Attention sinks are by definition tokens with mysteriously high attention score. Here, we first investigate the attention score difference between the secondary sinks and the BOS sink. 

% We then show that within secondary sinks, their exhibits different levels of sinks rates and they can last for variable number of layers. Both of these levels are correlated with the norm of the MLP output at which they get converted from.

%Attention sinks are tokens that, by definition, receive unusually high attention scores. 
In this section, we first examine the difference in attention scores between secondary sinks and primary sinks. We then show that secondary sinks exhibit varying sink-score and persist for different numbers of layers. Both the sink-score and the duration across layers are correlated with the norm of the MLP output from which the sinks are formed.

\paragraph{Attention Sink Score}
We follow~\cite{two-sides-of-the-same-coin} to define the attention sink-score for token position $k$ at layer $l$ as:
\begin{equation}
    \text{sink-score}_k^{(l,h)} = \frac{1}{T-k}\sum_{t=k}^{T-1}a_{tk}^{(l,h)}, \qquad \text{sink-score}_k^{l} = \frac{1}{H}\sum_{h=0}^{H} \text{sink-score}_k^{(l,h)}
\end{equation}

\begin{figure}[h!]
    \centering
    \includegraphics[width=\linewidth]{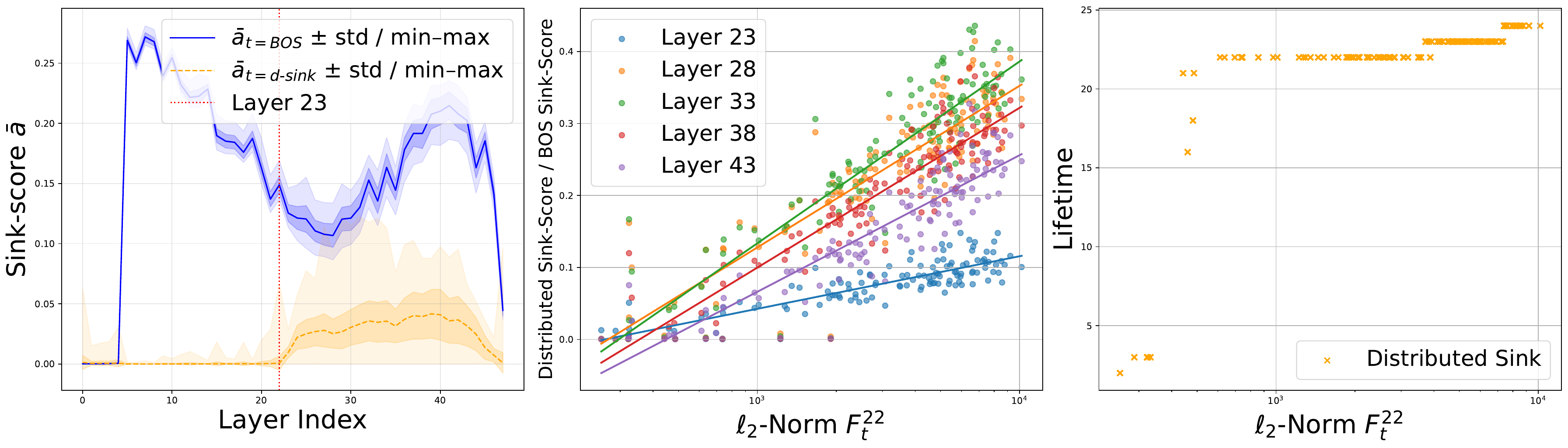}
    \caption{\textbf{(Left)} Sink-score comparison between BOS and \mymethod{} across whole model depth. \textbf{(Mid)} \mymethod{} Sink-score grows log-linearly with $\ell_2$-norm $l_\text{start}$. \textbf{(Right)} Sink Lifetime growths with $\ell_2$-norm $l_\text{start}$.}
    \label{fig:lifetime}
\end{figure}

As shown in
% ~\Cref{fig:bos sink-score}
~\Cref{fig:lifetime} (Left)
, the sink-score for the BOS sink shows a valley shape across the model depth, and it is when it reaches its lowest sink score that the secondary sinks started emerging. This suggests the secondary sinks may serve to compensate for the decay in the BOS sink. 
% \jf{TODO: add another model result figure here} \cz{increase font size in fig 7}

% \input{figures/fig_attention_weight_comparison}
% \begin{figure}[h!]
%     \centering
%     \includegraphics[width=0.9\linewidth]{figures/raws/norm_sink-score_lifetime.pdf}
%     \caption{Lifetime, sink rate vs creation norm}
%     \label{fig:lifetime}
% \end{figure}

\paragraph{Attention Sink Lifetime}
% We observe that secondary sinks can last for varying numbers of layers ranging from as small as 2 layers to 22 layers which is half of the network. To quantify this, we define the attention sink lifetime as the difference between the layer from which the sink vanished $l_{end}$ and the layer $l_{start}$ that the sink got created which is 22 for Deepseek-14B. ~\Cref{fig:lifttime} shows that both lifetime and the sink-score of the secondary sink are correlated highly with the $\ell_2$-norm of the MLP output $\vf_t^{22}$ at $l_{start}$. The ratio of Sink-socre between the secondary sink and BOS sink shows a linear relationship with largest slope at layer 33. Meanwhile, lifetime grows in a piecewise function with the log-norm.

We observe that secondary sinks persist for varying depths, ranging from as few as 2 layers to as many as 22 layers, approximately half of the network. 
% To quantify this behavior, we define the attention sink lifetime as:
% \[l_{\text{end}} - l_{\text{start}},\]
% where $l_{\text{start}}=22$ is the layer at which the secondary sink first emerges in DeepSeek-14B, and $l_{\text{end}}$ is the layer at which it vanishes. 
As shown in \Cref{fig:lifetime} (right), both the sink lifetime and the sink score of the secondary sink are strongly correlated with the $\ell_2$-norm of the MLP output $\vf^{22}_t$ at $l_{\text{start}}$. The ratio between the sink scores of the secondary sink and the BOS sink exhibits a log-linear relationship with $\vf^{22}_t$, with the largest slope observed at layer 33. In contrast, the sink lifetime increases monotonically with the log-norm, exhibiting distinct linear regimes separated by plateaus.

\section{Conclusion} 
% This work identifies a class of multi-token sinks that show an hierarchical levels within the model. We focus empirically on its properties, formation and effect on attention on well-trained LLMs. \addref{} has related attention sinks to the pre-training objective on BOS attention sinks and \addref{} has related multi-token sinks to positional embedding. Hierarchical sinks emerge with different position encoding during pre-training and how post-training porcess enhance the affect of hierarchical sinks are promising directions for next step. Additionally, how different levels of sinks related to reasoning generation remains unclear.

This work identifies a class of multi-token attention sinks that in certain LLMs, that differ from those described in previous works. We empirically study their properties, formation, and influence on attention patterns. 
% Prior work~\citep{When-attention-sink-emerges} links BOS attention sinks to the pre-training objective, while \citep{What-are-you-sinking} relates multi-token sinks to positional embeddings. 
We have presented evidence that secondary sinks are fundamentally different from BOS sinks. Investigating the root cause of secondary sinks emergence during pre-training, as well as why post-training processes amplify their effects, is a promising direction for future research. Moreover, its effect on the text generation and downstream performance remains an open question.

% under different positional encoding schemes during pre-training, as well as why post-training processes amplify their effects, is a promising direction for future research. x
% Moreover, its effect on the text generation and downstream performance remains an open question.

% how different levels of secondary sink affect reasoning trace generation remains an open question.

\newpage
\bibliography{ref}
\bibliographystyle{iclr2026_conference}

\newpage
\appendix
\section{Additional Detection Results}
\label{apx:sec:detection results}
\begin{table*}[ht]
\centering
\small
\scalebox{0.65}{
\begin{tabular}{llr|cccc}
\toprule
\multirow{2}{*}{Model Family} & \multirow{2}{*}{Model Name} & \multirow{2}{*}{RoPE Base $\theta$} & Primary & Primary & Top 3 Secondary & Secondary \\
& & & Sink Layer & Sink Count & Sink Layers & Sink Count \\
% \multirow{2}{*}{Model Family} & \multirow{2}{*}{Model Name} & \multirow{2}{*}{RoPE Base $\theta$} &
% \multicolumn{2}{c}{Primary Sink} & \multicolumn{2}{c}{Secondary Sink} \\
\midrule
DeepSeek
& deepseek-ai/DeepSeek-R1-Distill-Llama-8B & 500000     & 1	& 30    & X & X \\
& deepseek-ai/DeepSeek-R1-Distill-Qwen-1.5B & 10000     & 2	& 60    & X & X \\
& deepseek-ai/DeepSeek-R1-Distill-Qwen-7B & 10000       & 3	& 60    & 11: 82.55\% 12: 9.19\%                & 1284 \\
& deepseek-ai/DeepSeek-R1-Distill-Qwen-14B & 1000000    & 4	& 60    & 22: 100\%                             & 75 \\
& deepseek-ai/DeepSeek-R1-Distill-Qwen-32B & 1000000    & 4	& 90    & 38: 98.08\%, 39: 1.92\%               & 52 \\
\midrule
Qwen3
& Qwen/Qwen3-0.6B-Base & 1000000            & 2 & 30 & X & X\\
& Qwen/Qwen3-0.6B & 1000000                 & 2 & 30 & X & X\\
& Qwen/Qwen3-1.7B-Base & 1000000            & 2 & 30 & X & X\\
& Qwen/Qwen3-1.7B & 1000000                 & 2 & 30 & X & X\\
& Qwen/Qwen3-4B-Base & 1000000              & 6 & 30 & 16: 100\% & 86\\
& Qwen/Qwen3-4B & 1000000                   & 6 & 30 & 16: 100\% & 105\\
& Qwen/Qwen3-4B-Thinking-2507 & 5000000     & 6 & 30 & 16: 100\% & 81\\
& Qwen/Qwen3-4B-Instruct-2507 & 5000000     & 6 & 30 & 16: 100\% & 80\\
& Qwen/Qwen3-8B-Base & 1000000              & 6 & 30 & 16: 100\% & 104\\
& Qwen/Qwen3-8B & 1000000                   & 6 & 30 & 16: 100\% & 134\\
& Qwen/Qwen3-14B-Base & 1000000             & 6 & 30 & 9: 80.51\%, 8: 14.07\%, 10: 3.13\% & 6803\\
& Qwen/Qwen3-14B & 1000000                  & 6 & 30 & 9: 74.0\%, 8: 13.85\%, 38: 9.4\% & 7662\\
& Qwen/Qwen3-32B & 1000000                  & 6 & 30 & 9: 38.36\%, 8: 36.67\%, 7: 23.06\% & 33727\\
\midrule
Qwen2.5
& Qwen/Qwen2.5-1.5B & 1000000               & 2 & 30 & X & X\\
& Qwen/Qwen2.5-1.5B-Instruct & 1000000      & 2 & 30 & X & X\\
& Qwen/Qwen2.5-7B & 1000000                 & 3 & 30 & 11: 50\%, 12: 50\% & 32\\
& Qwen/Qwen2.5-7B-Instruct & 1000000        & 3 & 30 & 12: 56.76\%, 11: 43.24\% & 37\\
& Qwen/Qwen2.5-14B & 1000000                & 5 & 30 & 22: 100\%        & 87\\
& Qwen/Qwen2.5-14B-Instruct & 1000000       & 5 & 30 & 22: 100\%        & 94\\
& Qwen/Qwen2.5-32B & 1000000                & 5 & 30 & 38: 98.61\%, 40: 1.39\% & 72\\
& Qwen/Qwen2.5-32B-Instruct & 1000000       & 5 & 30 & 38: 100\% & 59\\
\midrule
Qwen2.5-Math
& Qwen/Qwen2.5-Math-1.5B & 10000            & 2 & 30 & 4: 100\% & 2777\\
& Qwen/Qwen2-Math-1.5B-Instruct & 10000     & 1 & 30 & 4: 100\% & 1622\\
& Qwen/Qwen2.5-Math-7B & 10000              & 3 & 30 & 11: 61.91\%, 12: 26.34\%, 13: 6.23\%        & 95854\\
& Qwen/Qwen2-Math-7B-Instruct & 10000       & 3 & 30 & 12: 35.68\%, 11: 29.37\%, 13: 17.78\%        & 13423\\
\midrule
Qwen2
& Qwen/Qwen2-1.5B & 1000000                 & 1 & 30 & X & X\\
& Qwen/Qwen2-1.5B-Instruct & 1000000        & 1 & 30 & X & X\\
& Qwen/Qwen2-7B & 1000000                   & 3 & 30 & 17: 42.86\%, 19: 14.29\%, 14: 14.29\% & 7\\
& Qwen/Qwen2-7B-Instruct & 1000000          & 3 & 30 & 18: 40\%, 16: 20\%, 19: 20\% & 10\\
\midrule
Qwen2-Math
& Qwen/Qwen2-Math-1.5B & 10000              & 3 & 30 & 3: 87.64\%, 4: 12.36\% & 356\\
& Qwen/Qwen2-Math-1.5B-Instruct & 10000     & 1 & 30 & 3: 100\% & 103\\
& Qwen/Qwen2-Math-7B & 10000                & 4 & 30 & 12: 33.16\%, 13: 21.8\%, 14: 14.94\% & 68709\\
& Qwen/Qwen2-Math-7B-Instruct & 10000       & 3 & 30 & 12: 35.53\%, 13: 22.88\%, 11: 15.3\% & 49682\\
\midrule
Llama-3.1
& meta-llama/Llama-3.1-8B & 500000 & 1 & 30 & X & X \\
& meta-llama/Llama-3.1-8B-Instruct & 500000 & 1 & 30 & X & X \\
\midrule
Phi-4
& microsoft/phi-4 & 250000                  & 2 & 30 & 4: 100\% & 30 \\
& microsoft/Phi-4-reasoning & 500000        & 2 & 30 & X \\
& microsoft/phi-4-mini-reasoning & 10000    & 3 & 30 & X \\
& microsoft/Phi-4-mini-instruct & 10000     & 3 & 30 & X \\
\midrule
Mathstral
& mistralai/Mathstral-7B-v0.1 & 1000000     & 1 & 30 & X & X \\
\midrule
Code-LLaMA
& meta-llama/CodeLlama-7b-hf & 1000000          & 1 & 30 & X & X \\
& codellama/CodeLlama-7b-Instruct-hf & 1000000  & 1 & 60 & X & X \\
& meta-llama/CodeLlama-13b-hf & 1000000         & 2 & 30 & X & X \\
& codellama/CodeLlama-13b-Instruct-hf & 1000000 & 2 & 30 & X & X \\
\midrule
QwQ
& Qwen/QwQ-32B & 1000000 & 1 & 30 & 38: 97.37\%, 40: 2.63\% & 38 \\
\bottomrule
\end{tabular}
}
\caption{Model configurations and secondary sink detection results based on 30 traces generated by DeepSeek-14B on AIME24. X indicates that no secondary sinks were detected. \textit{RoPE Base $\theta$} denotes the parameter used in computing the rotary frequency; larger $\theta$ means lower frequency. \textit{Primary Sink Layer} indicates the layer at which primary sinks appear. \textit{Top 3 Secondary Sink Layers} denotes the top 3 creation layer $l_{\text{start}}$ at which tokens are converted into secondary sinks, with $N\%$ indicating the proportion of total sinks created by that layer. \textit{Secondary Sink Count} reports the total number of detected secondary sink tokens.}
\label{tab: detection results}
\end{table*}

\begin{table*}[ht]
\centering
\small
\scalebox{0.65}{
\begin{tabular}{llr|cccc}
\toprule
\multirow{2}{*}{Model Family} & \multirow{2}{*}{Model Name} & \multirow{2}{*}{RoPE Base $\theta$} & Primary & Primary & Top 3 Secondary & Secondary \\
& & & Sink Layer & Sink Count & Sink Layers & Sink Count \\
% \multirow{2}{*}{Model Family} & \multirow{2}{*}{Model Name} & \multirow{2}{*}{RoPE Base $\theta$} &
% \multicolumn{2}{c}{Primary Sink} & \multicolumn{2}{c}{Secondary Sink} \\
\midrule
DeepSeek
& deepseek-ai/DeepSeek-R1-Distill-Llama-8B & 500000     & 1	& 500     &X    & X \\
& deepseek-ai/DeepSeek-R1-Distill-Qwen-1.5B & 10000     & 2	& 1000    &X    & X \\
& deepseek-ai/DeepSeek-R1-Distill-Qwen-7B & 10000       & 3	& 1000    &11: 69.14\%, 4: 22.4\%, 12: 6.39\%  & 5807 \\
& deepseek-ai/DeepSeek-R1-Distill-Qwen-14B & 1000000    & 4	& 1500    &22: 99.53\%, 23: 0.47\%    & 214 \\
& deepseek-ai/DeepSeek-R1-Distill-Qwen-32B & 1000000    & 4	& 1000    &38: 98.55\%, 39: 1.21\%; 40: 0.24\% & 413 \\
\midrule
Qwen3
& Qwen/Qwen3-0.6B-Base & 1000000            & 2 & 500 & X& X        \\
& Qwen/Qwen3-0.6B & 1000000                 & 2 & 500 & X& X    \\
& Qwen/Qwen3-1.7B-Base & 1000000            & 2 & 500 & X& X    \\
& Qwen/Qwen3-1.7B & 1000000                 & 2 & 500 & X& X    \\
& Qwen/Qwen3-4B-Base & 1000000              & 6 & 500 & 16: 100\%& 504    \\
& Qwen/Qwen3-4B & 1000000                   & 6 & 500 & 16: 100\%& 692    \\
& Qwen/Qwen3-4B-Thinking-2507 & 5000000     & 6 & 500 & 16: 100\%& 645    \\
& Qwen/Qwen3-4B-Instruct-2507 & 5000000     & 6 & 500 & 16: 100\%& 623    \\
& Qwen/Qwen3-8B-Base & 1000000              & 6 & 500 & 16: 100\%& 684    \\
& Qwen/Qwen3-8B & 1000000                   & 6 & 500 & 16: 100\%& 829\\
& Qwen/Qwen3-14B-Base & 1000000             & 6 & 500 & 9: 84.96\%, 8: 9.37\%, 10: 3.5\%& 36533\\
& Qwen/Qwen3-14B & 1000000                  & 6 & 500 & 9: 81.97\%, 8: 9.07\%, 38: 5.56\%& 37349\\
& Qwen/Qwen3-32B & 1000000                  & 6 & 500 & 8: 37.0\%, 9: 34.92\%, 7: 26.68\%& 203789\\
\midrule
Qwen2.5
& Qwen/Qwen2.5-1.5B & 1000000               & 2 & 500 & X& X\\
& Qwen/Qwen2.5-1.5B-Instruct & 1000000      & 2 & 500 & X& X\\
& Qwen/Qwen2.5-7B & 1000000                 & 3 & 500 & 11: 48.44\%, 12: 48.44\%, 13: 1.56\%& 64\\
& Qwen/Qwen2.5-7B-Instruct & 1000000        & 3 & 500 & 12: 54.17\%, 11: 41.67\%, 15: 4.17\%& 72\\
& Qwen/Qwen2.5-14B & 1000000                & 5 & 500 & 22: 99.61\%, 23: 0.39\%& 258\\
& Qwen/Qwen2.5-14B-Instruct & 1000000       & 5 & 500 & 22: 100\%& 222\\
& Qwen/Qwen2.5-32B & 1000000                & 5 & 500 & 38: 98.39\%, 39: 1.41\%, 40: 0.2\%& 497\\
& Qwen/Qwen2.5-32B-Instruct & 1000000       & 5 & 500 & 38: 99.35\%, 39: 0.65\%& 309\\
\midrule
Qwen2.5-Math
& Qwen/Qwen2.5-Math-1.5B & 10000            & 2 & 500 & 4: 100\%& 6667\\
& Qwen/Qwen2-Math-1.5B-Instruct & 10000     & 1 & 500 & 4: 98.74\%, 1: 0.87\%, 2: 0.38\%& 4700\\
& Qwen/Qwen2.5-Math-7B & 10000              & 3 & 500 & 11: 55.88\%, 12: 28.87\%, 13: 7.91\%& 149922\\
& Qwen/Qwen2-Math-7B-Instruct & 10000       & 3 & 500 & 12: 34.07\%, 11: 29.45\%, 13: 19.03\%& 24575\\
\midrule
Qwen2
& Qwen/Qwen2-1.5B & 1000000                 & 1 & 500 & X & X\\
& Qwen/Qwen2-1.5B-Instruct & 1000000        & 1 & 500 & X & X\\
& Qwen/Qwen2-7B & 1000000                   & 3 & 500 & 17: 30.0\%, 18: 30.0\%, 16: 15.0\% & 20\\
& Qwen/Qwen2-7B-Instruct & 1000000          & 3 & 500 & 17: 30.0\%, 18: 30.0\%, 15: 15.0\% & 20\\
\midrule
Qwen2-Math
& Qwen/Qwen2-Math-1.5B & 10000              & 3 & 500 & 3: 93.04\%, 4: 6.9\%, 6: 0.06\% & 1739\\
& Qwen/Qwen2-Math-1.5B-Instruct & 10000     & 1 & 500 & 3: 99.92\%, 4: 0.08\%  & 1238\\
& Qwen/Qwen2-Math-7B & 10000                & 4 & 500 & 12: 30.88\%, 13: 23.25\%, 14: 16.26\%   & 114579\\
& Qwen/Qwen2-Math-7B-Instruct & 10000       & 3 & 500 & 12: 33.96\%, 13: 22.37\%, 11: 17.11\%   & 82327\\
\midrule
Llama-3.1
& meta-llama/Llama-3.1-8B & 500000 & 1 & 500 & X & X \\
& meta-llama/Llama-3.1-8B-Instruct & 500000 & 1 & 500 & X & X \\
\midrule
Phi-4
& microsoft/phi-4 & 250000                  & 2 & 500 & 4: 100\% & 500 \\
& microsoft/Phi-4-reasoning & 500000        & 2 & 500 & X \\
& microsoft/phi-4-mini-reasoning & 10000    & 3 & 500 & X \\
& microsoft/Phi-4-mini-instruct & 10000     & 3 & 500 & X \\
\midrule
Mathstral
& mistralai/Mathstral-7B-v0.1 & 1000000     & 1 & 500 & X & X \\
\midrule
Code-LLaMA
& meta-llama/CodeLlama-7b-hf & 1000000          & 1 & 1000 & X & X \\
& codellama/CodeLlama-7b-Instruct-hf & 1000000  & 1 & 1000 & X & X \\
& meta-llama/CodeLlama-13b-hf & 1000000         & 2 & 500 & X & X \\
& codellama/CodeLlama-13b-Instruct-hf & 1000000 & 2 & 500 & X & X \\
\midrule
QwQ
& Qwen/QwQ-32B & 1000000                        & 1 & 500 & 38: 95.69\%, 40: 2.59\%, 39: 0.86\% & 116 \\
\bottomrule
\end{tabular}
}
\caption{Model configurations and secondary sink detection results based on 500 traces generated by DeepSeek-14B on AIME-500. X indicates that no secondary sinks were detected. \textit{RoPE Base $\theta$} denotes the parameter used in computing the rotary frequency; larger $\theta$ means lower frequency. \textit{Primary Sink Layer} indicates the layer at which primary sinks appear. \textit{Top 3 Secondary Sink Layers} denotes the top 3 creation layer $l_{\text{start}}$ at which tokens are converted into secondary sinks, with $N\%$ indicating the proportion of total sinks created by that layer. \textit{Secondary Sink Count} reports the total number of detected secondary sink tokens.}
\label{tab: detection results}
\end{table*}

% \section{Sink Levels across models}
% \jf{todo add more results here}
% \section{Token Swapping}
% \paragraph{Token Swapping}
% \label{apx:sec:token swapping}
\newpage
\section{Properties of Secondary Sinks}

\begin{figure}[h!]
    \centering
    \begin{minipage}{0.48\linewidth}
        \centering
        \includegraphics[width=\linewidth]{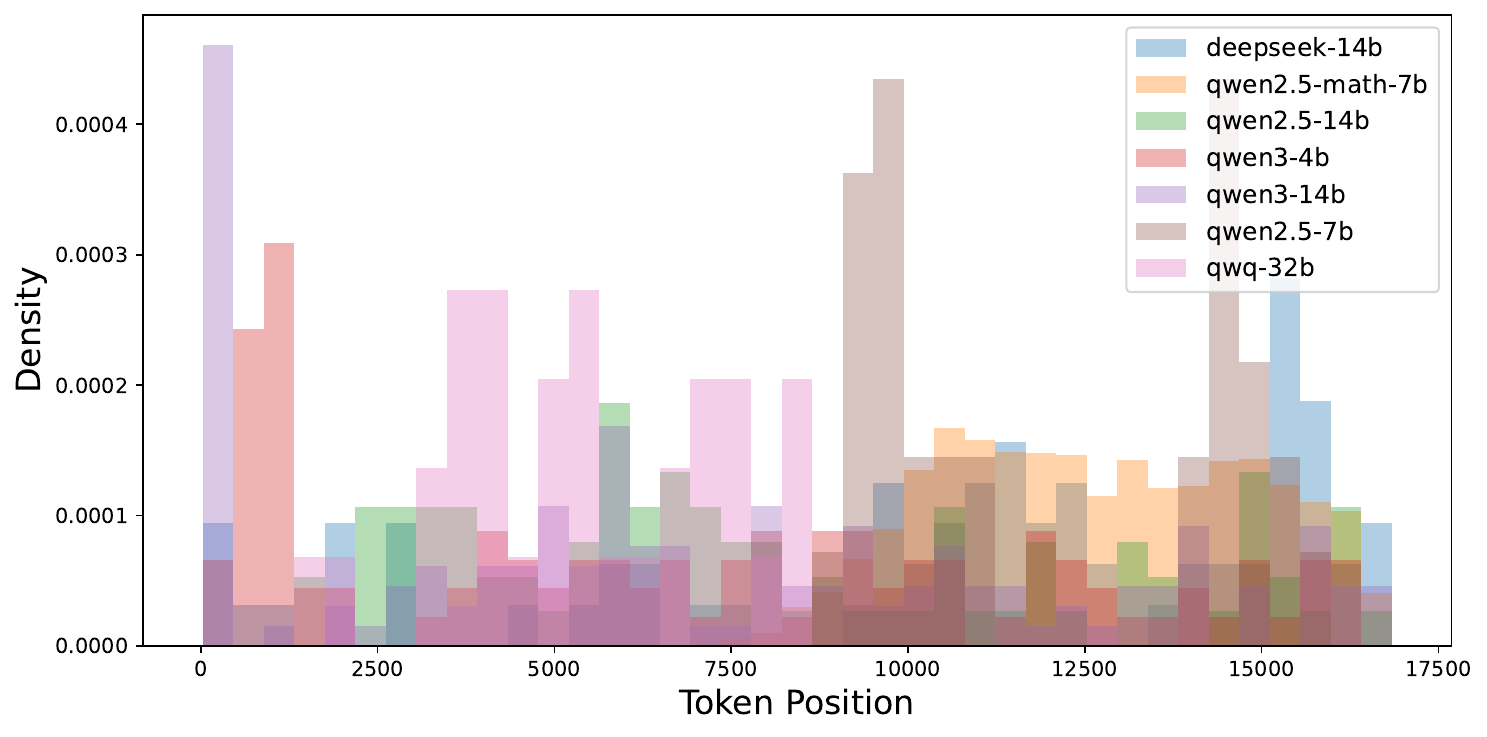}
        \caption{Distribution of sinks across token index.}
        \label{fig:token-pos-density}
    \end{minipage}\hfill
    \begin{minipage}{0.5\linewidth}
        \centering
        \vspace{1mm}
        \resizebox{\linewidth}{!}{
            \begin{tabular}{l | c c c c c c c}
                \toprule
                \textbf{Model} & \multicolumn{7}{c}{\textbf{\mymethod{} Token Distribution}} \\
                \midrule
                \multirow{2}{*}{Deepseek-14b}
                & \texttt{" "} & \texttt{"1"} & \texttt{"\^{ }"} & \texttt{"2"}
                & \texttt{","} & \texttt{"("} & \texttt{" I"} \\
                & 56.41\% & 17.95\% & 5.13\% & 5.13\% & 2.56\% & 2.56\% & 2.56\% \\[4pt]

                \multirow{2}{*}{Qwen3-14b}
                & \texttt{"\textbackslash n"} & \texttt{" "} & \texttt{"1"} & \texttt{"0"}
                & \texttt{"9"} & \texttt{"2"} & \texttt{","} \\
                & 19.87\% & 19.87\% & 13.91\% & 9.93\% & 4.64\% & 3.97\% & 3.97\% \\[4pt]

                \multirow{2}{*}{Qwen2-7B}
                & \texttt{","} & \texttt{"1"} & \texttt{"2"} & \texttt{" "}
                & \texttt{".\textbackslash n"} & \texttt{"0"} & \texttt{"4"} \\
                & 6.13\% & 5.28\% & 3.62\% & 3.56\% & 3.50\% & 3.48\% & 2.98\% \\[4pt]

                \multirow{2}{*}{Qwen2.5-14B}
                & \texttt{" "} & \texttt{"0"} & \texttt{"1"} & \texttt{"8"}
                & \texttt{"="} & \texttt{"\_"} & \texttt{"t"} \\
                & 74.71\% & 9.20\% & 5.75\% & 2.30\% & 2.30\% & 1.15\% & 1.15\% \\
                \bottomrule
            \end{tabular}
        }
        \vspace{4mm}
        \captionof{table}{Frequency of secondary sink tokens.}
        \label{tab:token-distribution}
    \end{minipage}
\end{figure}

\section{Causal formation of Secondary Sinks}
\begin{figure}[h!]
    \centering
    \includegraphics[width=0.8\linewidth]{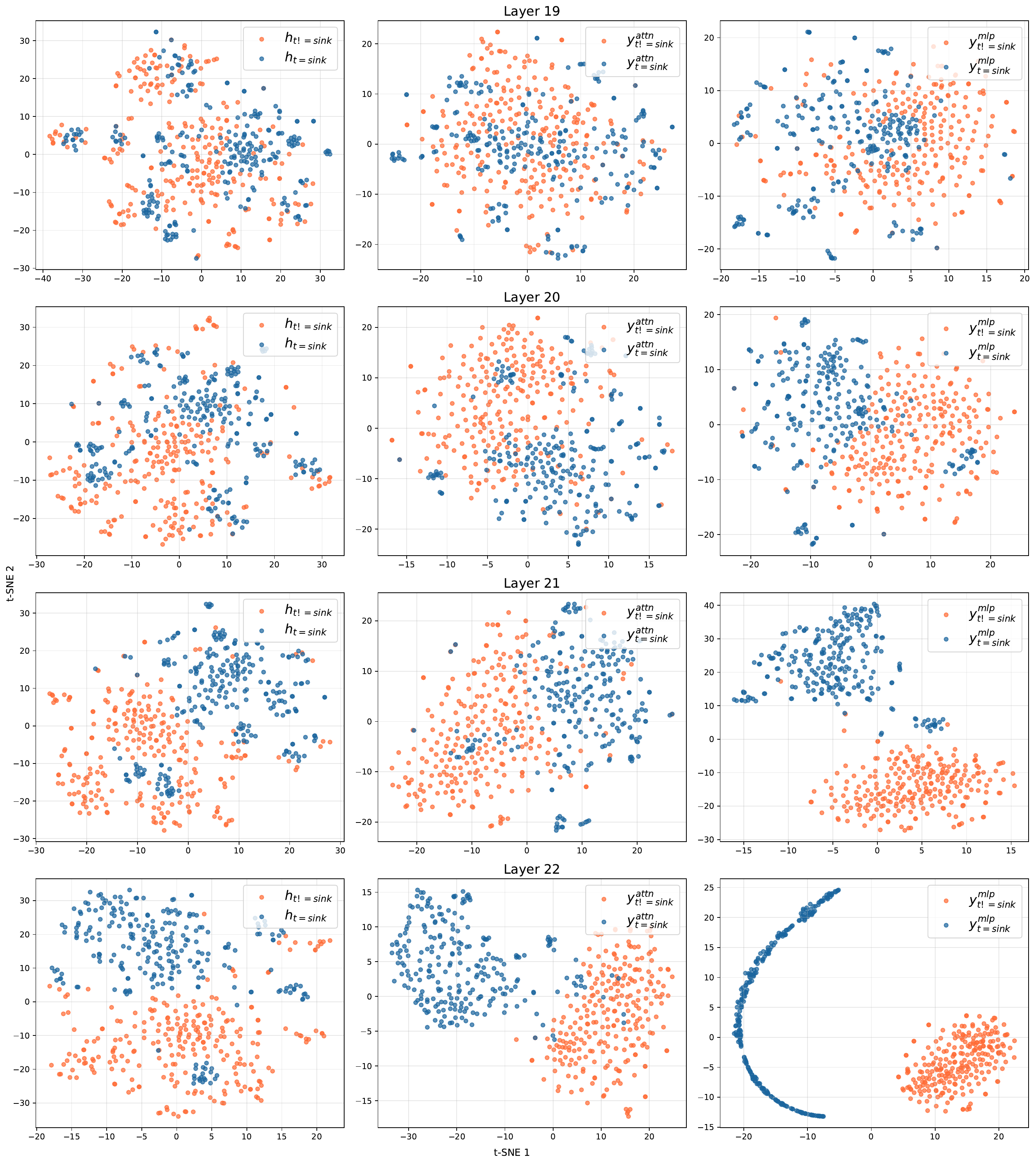}
    \caption{Hidden State, Attention Output and MLP Output t-sne Clustering between normal semantic uninformative tokens and future sinks tokens.}
    \label{fig:clustering}
\end{figure}

% \begin{figure}[h!]
%     \centering
%     \begin{subfigure}{0.25\linewidth}
%         \centering
%         \includegraphics[width=\linewidth]{figures/raws/decoder_overview.pdf}
%         \caption{Decoder overview}
%     \end{subfigure}\hfill
%     \begin{subfigure}{0.6\linewidth}
%         \centering
%         \includegraphics[width=\linewidth]{figures/raws/tsne_plots.pdf}
%         \caption{Activation swapping}
%     \end{subfigure}
%     \caption{Swapping activation at attention output $a_t$, residual input $r_t$, MLP input $x_t^{mlp}$, and MLP output $y_t^{mlp}$.}
%     \label{fig:activation_swap}
% \end{figure}

\begin{figure}[h!]
    \centering
    \includegraphics[width=0.82\linewidth]{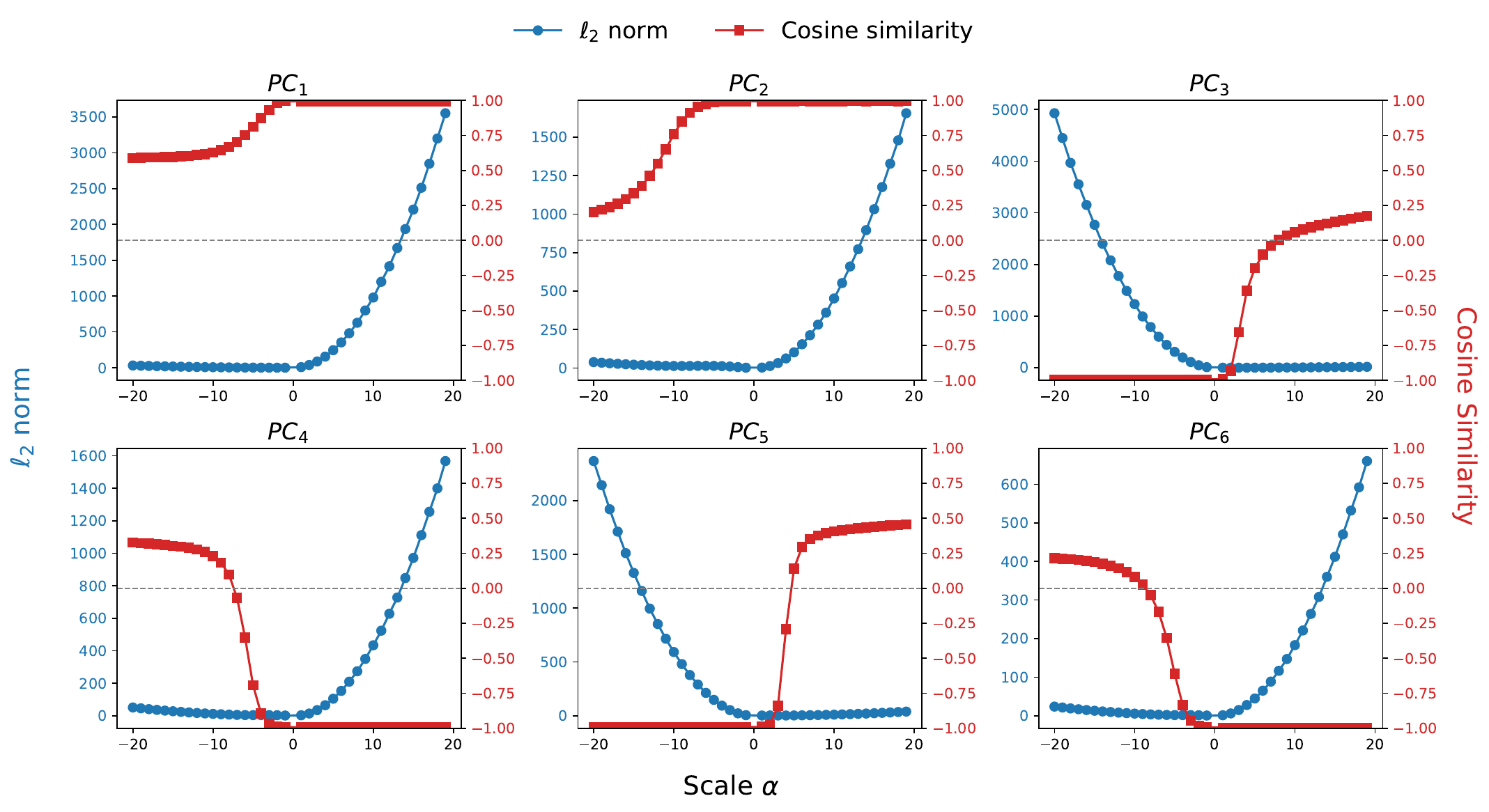}
    \caption{The $\ell_2$-norm and cosine similarity of the MLP output relative to BOS sink with $\pm\alpha\mathrm{PC_i}$ as input.
    }
    \label{fig:pca}
\end{figure}

\begin{figure}[h!]
    \centering
    \includegraphics[width=0.8\linewidth]{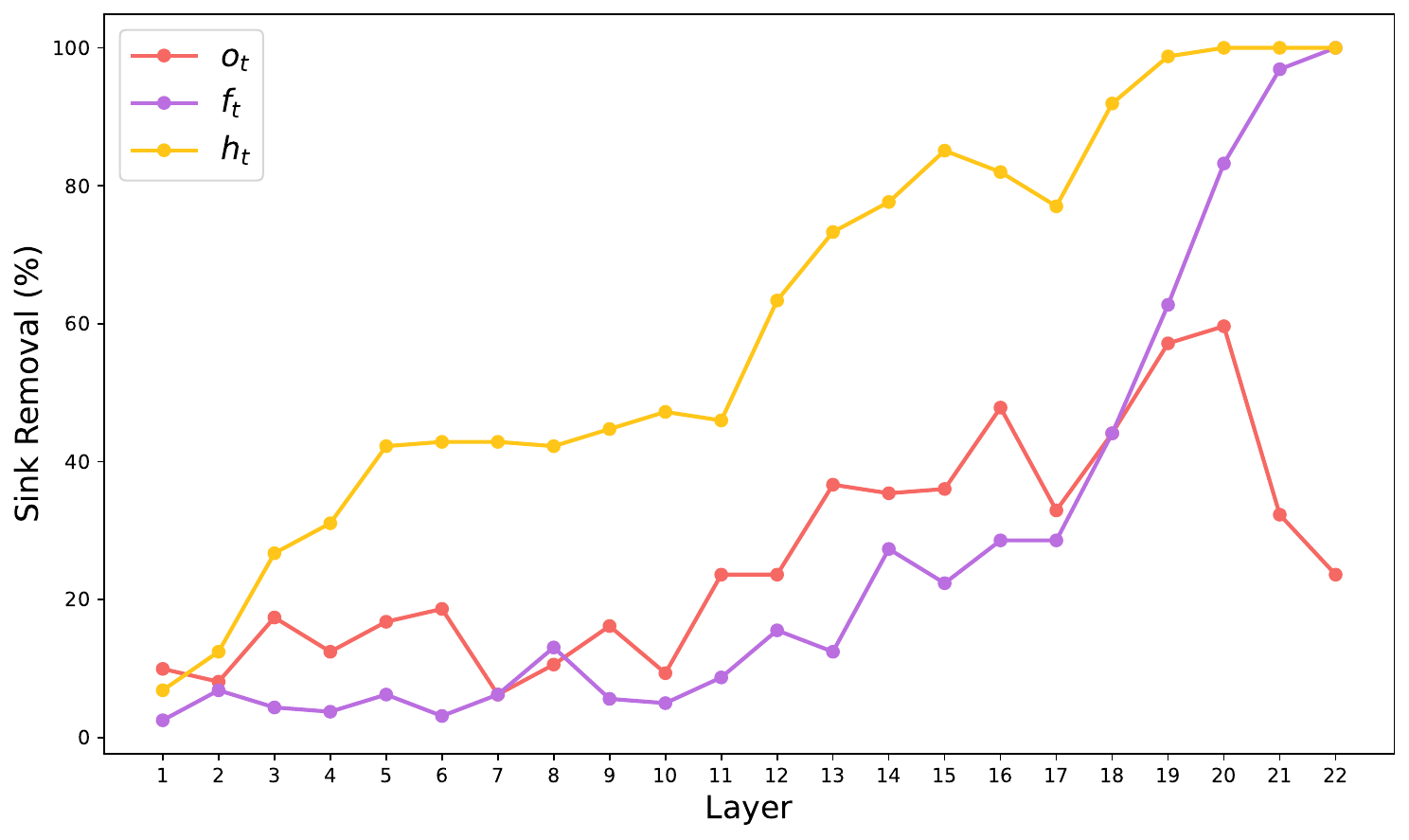}
    \caption{Swapping the activation of the secondary sink at hidden state $\vh_t^l$, attention output $\vo_t^l$, and MLP output $\vf_t^l$ with that of an average uninformative token.}
    \label{fig:activation_swap}
\end{figure}

% \subsection{Token Swapping}
\paragraph{Token Swapping}
\label{token-swapping}
To further verify the formation of secondary sinks in early layers,
we conduct token swapping experiments between future secondary sinks and average tokens\footnote{We average the hidden vectors of all non-sink tokens at a given layer to obtain the average token representation.}.
By swapping the hidden vectors of these tokens, MLP outputs, and attention outputs at early layers,
we observe that we can often suppress the sink from emerging in later layers. The later we swap out the activations, the more effective it is at suppressing the future sink. Figure \ref{fig:activation_swap} shows the percentage of the time the future sink is suppressed, as a function of the layer in which it is swapped.

% \begin{figure}[h!]
%     \centering
%     \begin{subfigure}{0.25\linewidth}
%         \centering
%         \includegraphics[width=\linewidth]{figures/raws/decoder_overview.pdf}
%         \caption{Decoder overview}
%     \end{subfigure}\hfill
%     \begin{subfigure}{0.7\linewidth}
%         \centering
%         \includegraphics[width=\linewidth]{figures/raws/activation_swap.pdf}
%         \caption{Activation swapping}
%     \end{subfigure}
%     \caption{Swapping activation at attention output $a_t$, residual input $r_t$, MLP input $x_t^{mlp}$, and MLP output $y_t^{mlp}$.}
%     \label{fig:activation_swap}
% \end{figure}

% To further verify the formation of secondary sinks in early layers,
% we conduct token swapping experiments between future secondary sinks and average tokens\footnote{We average the hidden vectors of all non-sink tokens at a given layer to obtain the average token representation.}.
% By swapping the hidden vectors of these tokens, MLP outputs, and attention outputs at early layers,
% we observe that we can often suppress the sink from emerging in later layers. The later we swap out the activations, the more effective it is at suppressing the future sink. Figure \ref{fig:activation_swap} shows the percentage of the time the future sink is suppressed, as a function of the layer in which it is swapped.

\end{document}